\documentclass[%
 reprint,
 amsmath,amssymb,
 aps, 
 pre,
 showkeys
]{revtex4-1}

\usepackage{graphicx}
\usepackage{dcolumn}
\usepackage{bm}

\usepackage{my-style}
\graphicspath{{figs/}}
\usepackage{algpseudocode, algorithm}
\usepackage{multirow}

\allowdisplaybreaks[4]

\begin{document}

\preprint{APS/123-QED}

\title{Empirical Bayes Method for Boltzmann Machines}

\author{Muneki Yasuda}  
\email{muneki@yz.yamagata-u.ac.jp}
\affiliation{%
 Graduate School of Science and Engineering, Yamagata University, Japan.
}%
\author{Tomoyuki Obuchi}  
\affiliation{%
 Department of Mathematical and Computing Science, Tokyo Institute of Technology, Japan.
}%


\begin{abstract}
In this study, we consider an empirical Bayes method for Boltzmann machines and propose an algorithm for it. 
The empirical Bayes method allows estimation of the values of the hyperparameters of the Boltzmann machine 
by maximizing a specific likelihood function referred to as the empirical Bayes likelihood function in this study. 
However, the maximization is computationally hard because the empirical Bayes likelihood function involves  
intractable integrations of the partition function. 
The proposed algorithm avoids this computational problem by using the replica method and the Plefka expansion. 
Our method does not require any iterative procedures and is quite simple and fast, 
though it introduces a bias to the estimate, which exhibits an unnatural behavior with respect to the size of the dataset. 
This peculiar behavior is supposed to be due to the approximate treatment by the Plefka expansion. 
A possible extension to overcome this behavior is also discussed. 
\end{abstract}

\pacs{Valid PACS appear here}
\maketitle


\section{Introduction} \label{sec:intro}

\textit{Boltzmann machine learning} (BML)~\cite{Ackley_etal1985} has been actively studied in the field of machine learning 
and also in statistical mechanics. 
In statistical mechanics, the problem of BML is sometimes referred to as the \textit{inverse Ising problem}, 
because a Boltzmann machine is the same as an Ising model, and BML can be regarded as an inverse problem for the Ising model.
The framework of the \textit{usual} BML is as follows.
Given a set of observed data points (e.g., spin snapshots), 
we estimate appropriate values of the parameters, the external field and couplings, of our Boltzmann machine through maximum likelihood (ML) estimation (cf. Sec.~\ref{sec:BM}). 
Because BML involves intractable multiple summations (i.e., evaluation of the partition function), 
many approximations for it were proposed from the viewpoint of statistical mechanics~\cite{Roudi2009}:
for example, methods based on mean-field approximations (such as the Plefka expansion~\cite{Plefka1982} and the cluster variation method~\cite{CVM-review2005})~\cite{Kappen1998,TTanaka1998,Yasuda2006,Monasson2009,Yasuda&Tanaka2009,Federico2012,Cyril2013}  
and methods based on other approximations~\cite{MPF2011,SMCI2015}. 

\begin{figure}[tb]
\centering
\includegraphics[height=3.4cm]{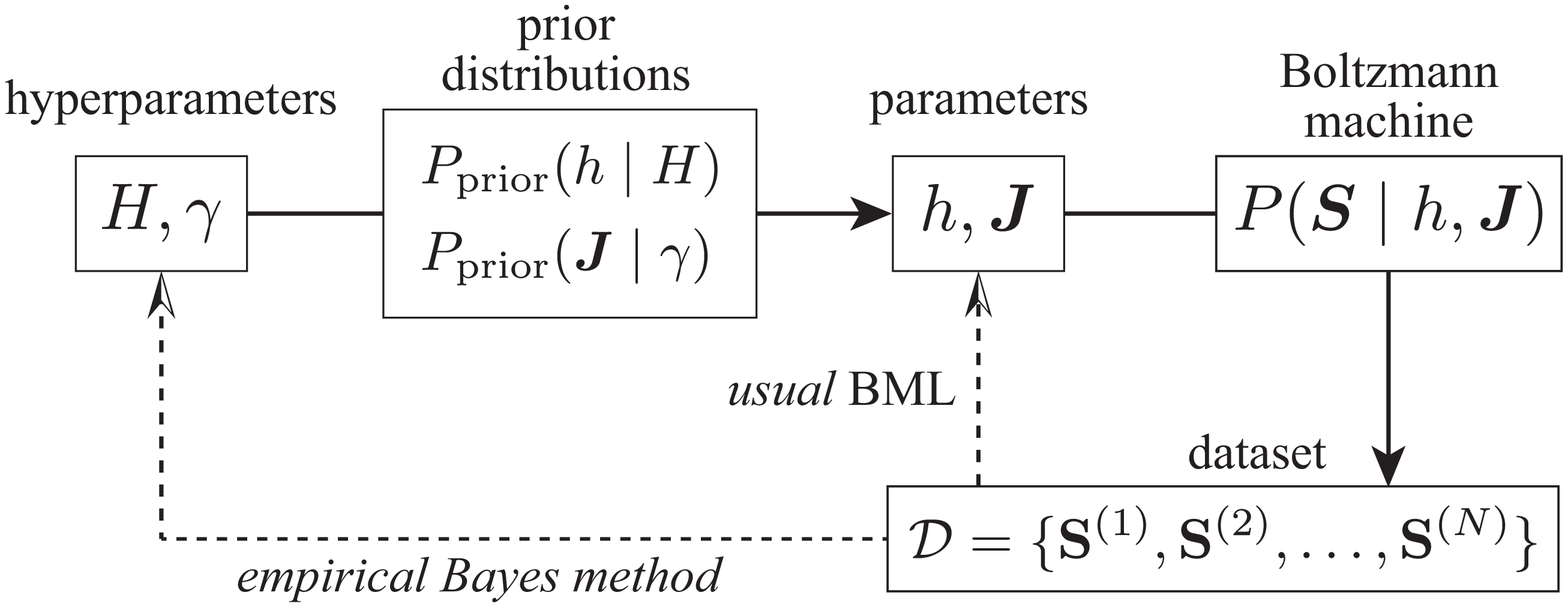}
\caption{Illustration of scheme of empirical Bayes method considered in this study.}
\label{fig:Scheme_of_EBM}
\end{figure}

In this study, we focus on another type of learning problem. 
We consider prior distributions of parameters of the Boltzmann machine 
and assume that the prior distributions are governed by some hyperparameters. 
The introduction of the prior distributions is strongly connected with the regularized ML estimation (cf. Sec.~\ref{sec:BM}).
As mentioned above, the aim of the \textit{usual} BML is to optimize the values of the parameters of the Boltzmann machine 
by using a set of observed data points. 
Meanwhile, the aim of the problem investigated in this study is the estimation of appropriate values of the hyperparameters from the dataset 
without estimating specific values of the parameters. 
One way to allow us to accomplish this from the Bayesian point of view is the \textit{empirical Bayes method} 
(or also called type-II ML estimation or evidence approximation)~\cite{MacKay1992,Bishop2006} (cf. Sec.~\ref{sec:Framework_EB}). 
The schemes of the \textit{usual} BML and of our problem are illustrated in Fig.~\ref{fig:Scheme_of_EBM}.

However, the evaluation of the likelihood function in the empirical Bayes method is again intractable,  
because it involves intractable multiple integrations of the partition function.
In this study, we analyze the empirical Bayes method for fully-connected Boltzmann machines, 
using statistical mechanical techniques based on the replica method~\cite{ParisiBook1987,Nishimori2001} and the Plefka expansion 
to derive an algorithm for it. 
We consider two types of cases of the prior distribution of $\bm{J}$: the cases of Gaussian and Laplace priors. 

The rest of this paper is organized as follows.
The formulations of the \textit{usual} BML and the empirical Bayes method are presented in Sec.~\ref{sec:BM&EB}. 
In Sec.~\ref{sec:StatisticalMechanicalAnalysis}, we describe our statistical mechanical analysis for the empirical Bayes method. 
The proposed inference algorithm obtained from our analysis is shown in Sec.~\ref{sec:algorithm} with its pseudocode. 
In Sec.~\ref{sec:experiment}, we examine our proposed method through numerical experiments. 
Finally, the summary and some discussions are presented in Sec.~\ref{sec:summary}.

\section{Boltzmann Machine and Empirical Bayes Method}
\label{sec:BM&EB}

\subsection{Boltzmann machine and prior distributions}
\label{sec:BM}

Consider a fully-connected Boltzmann machine with $n$ Ising variables $\bm{S}:= \{S_i \in \{-1,+1\} \mid i = 1,2,\ldots, n\}$~\cite{Ackley_etal1985}:  
\begin{align} 
P(\bm{S} \mid h,\bm{J}):=\frac{1}{Z(h,\bm{J})}\exp\Big(h \sum_{i=1}^n S_i + \sum_{i<j}J_{ij}S_iS_j\Big),
\label{eqn:BoltzmannMachine}
\end{align}
where $\sum_{i<j}$ is the sum over all the distinct pairs of variables; i.e., $\sum_{i<j} = \sum_{i=1}^n\sum_{j = i+1}^n$. 
$Z(h,\bm{J})$ is the partition function defined by
\begin{align*}
Z(h,\bm{J}):= \sum_{\bm{S}}\exp\Big(h \sum_{i=1}^n S_i + \sum_{i<j}J_{ij}S_iS_j\Big),
\end{align*}
where $\sum_{\bm{S}}$ is the sum over all the possible configurations of $\bm{S}$; i.e., $\sum_{\bm{S}} := \prod_{i=1}^n \sum_{S_i = \pm 1}$.
The parameters, $h \in (-\infty, +\infty)$ and $\bm{J} := \{J_{ij} \in (-\infty, +\infty) \mid i<j\}$, denote the external field and couplings, respectively. 

Given $N$ observed data points, $\mcal{D}:=\{\mbf{S}^{(\mu)} \in \{-1,+1\}^n \mid \mu = 1,2,\ldots, N\}$, 
we define the log-likelihood function: 
\begin{align}
L_{\mrm{ML}}(h,\bm{J}):=\frac{1}{n N}\sum_{\mu = 1}^N \ln P(\mbf{S}^{(\mu)} \mid h,\bm{J}).
\label{eqn:log-likelihood}
\end{align}
Maximizing the log-likelihood function with respect to $h$ and $\bm{J}$ (i.e., the ML estimation) 
just corresponds to the BML (or the inverse Ising problem), 
i.e., 
\begin{align}
\{\hat{h}_{\mrm{ML}},\hat{\bm{J}}_{\mrm{ML}}\} = \argmax_{h, \bm{J}}L_{\mrm{ML}}(h,\bm{J}).
\label{eqn;InverseIsing}
\end{align} 

Now, we introduce prior distributions for the parameters $h$ and $\bm{J}$ as 
$P_{\mrm{prior}}(h\mid H)$ and
\begin{align}
P_{\mrm{prior}}(\bm{J} \mid \gamma)&:= \prod_{i<j} P_{\mrm{prior}}(J_{ij} \mid \gamma),
\label{eqn:prior_J}
\end{align}
respectively. 
$H$ and $\gamma$ are the hyperparameters of these prior distributions. 
One of the most important motivations for introducing the prior distributions is 
for a Bayesian interpretation of the regularized ML estimation~\cite{Bishop2006}. 
Given the observed dataset $\mcal{D}$, by using the prior distributions, 
the posterior distribution of $h$ and $\bm{J}$ is expressed as
\begin{align}
&P_{\mrm{post}}(h,\bm{J} \mid \mcal{D}, H, \gamma) \nn
&= \frac{P(\mcal{D} \mid h, \bm{J})P_{\mrm{prior}}(h\mid H)P_{\mrm{prior}}(\bm{J} \mid \gamma)}{P(\mcal{D} \mid H, \gamma)},
\label{eqn;posterior_H&J}
\end{align}
where
\begin{align*}
P(\mcal{D} \mid h, \bm{J}):= \prod_{\mu = 1}^N P(\mbf{S}^{(\mu)} \mid h,\bm{J}).
\end{align*}
The distribution in the denominator in Eq.~(\ref{eqn;posterior_H&J}), $P(\mcal{D} \mid H, \gamma)$, is sometimes referred to as the evidence. 
By using the posterior distribution, the maximum a posteriori (MAP) estimation of the parameters is obtained as
\begin{align}
\{\hat{h}_{\mrm{MAP}},\hat{\bm{J}}_{\mrm{MAP}}\} = \argmax_{h, \bm{J}}L_{\mrm{MAP}}(h,\bm{J}),
\label{eqn;MAP}
\end{align} 
where
\begin{align}
&L_{\mrm{MAP}}(h,\bm{J}):= \frac{1}{nN}\ln P_{\mrm{post}}(h,\bm{J} \mid \mcal{D}, H, \gamma)\nn
&=L_{\mrm{ML}}(h,\bm{J}) + \frac{1}{nN} R_0(h) + \frac{1}{nN} R_1(\bm{J}) + \mrm{constant}.
\end{align}
The MAP estimation in Eq.~(\ref{eqn;MAP}) corresponds to the regularized ML estimation, 
in which $R_0(h):=\ln P_{\mrm{prior}}(h\mid H)$ and $R_1(\bm{J}):=\ln P_{\mrm{prior}}(\bm{J} \mid \gamma)$ work as a penalty.
For example, (i) when the prior distribution of $\bm{J}$ is the Gaussian prior, 
\begin{align}
P_{\mrm{prior}}(J_{ij} \mid \gamma)= \sqrt{\frac{n}{2 \pi \gamma}} \exp\Big(-\frac{n J_{ij}^2}{2 \gamma}\Big),\quad \gamma > 0,
\label{eqn:GaussPrior}
\end{align}
$R_1(\bm{J})$ corresponds to the $L_2$ regularization term, and $\gamma$ corresponds to its coefficient; 
(ii) when the prior distribution of $\bm{J}$ is the Laplace prior, 
\begin{align}
P_{\mrm{prior}}(J_{ij} \mid \gamma)= \sqrt{\frac{n}{2 \gamma}} \exp\Big(-\sqrt{\frac{2n}{\gamma}} |J_{ij}|\Big),\quad \gamma > 0
\label{eqn:LaplacePrior}
\end{align}
$R_1(\bm{J})$ corresponds to the $L_1$ regularization term, and $\gamma$ again corresponds to its coefficient. 
The variances of these prior distributions are identical, $\mrm{Var}[J_{ij}]=\gamma /n$.
In this study, as a simple test case, we use these two prior distributions for $\bm{J}$ and  
\begin{align}
P_{\mrm{prior}}(h\mid H) = \delta(h - H),
\label{eqn:prior_H}
\end{align}
where $\delta(x)$ is the Dirac delta function, for $h$.

\subsection{Framework of the empirical Bayes method}
\label{sec:Framework_EB}

Using the empirical Bayes method, we can infer the values of the hyperparameters, $H$ and $\gamma$, from the observed dataset $\mcal{D}$.
We define a marginal log-likelihood function as
\begin{align}
L_{\mrm{EB}}(H,\gamma)&:=\frac{1}{nN} \ln \big[ P(\mcal{D} \mid h, \bm{J})\big]_{h,\bm{J}},
\label{eqn:EmpiricalBayesLikelihood}
\end{align}
where $[\cdots]_{h,\bm{J}}$ is the average over the prior distributions; i.e., 
\begin{align*}
[\cdots]_{h,\bm{J}}:= \int d\bm{J}\int d h (\cdots) P_{\mrm{prior}}(h\mid H)P_{\mrm{prior}}(\bm{J} \mid \gamma).
\end{align*}
We refer to the marginal log-likelihood function as the \textit{empirical Bayes likelihood function} in this study.
From the perspective of the empirical Bayes method, the optimal values of the hyperparameters, $\hat{H}$ and $\hat{\gamma}$, are obtained 
by maximizing of the empirical Bayes likelihood function; i.e., 
\begin{align}
\{\hat{H},\hat{\gamma}\} = \argmax_{H, \gamma} L_{\mrm{EB}}(H,\gamma).
\label{eqn:Maximizing_EBL}
\end{align}
It is noteworthy that $[P(\mcal{D} \mid h, \bm{J})]_{h,\bm{J}}$ in Eq.~(\ref{eqn:EmpiricalBayesLikelihood}) 
is identified as the evidence appearing in Eq.~(\ref{eqn;posterior_H&J}).

The marginal log-likelihood function can be rewritten as
\begin{align}
L_{\mrm{EB}}(H,\gamma)=\frac{1}{nN}\ln \Big[\exp\big(n N L_{\mrm{ML}}(h,\bm{J})\big)\Big]_{h, \bm{J}}
\label{eqn:EmpiricalBayesLikelihood_ML-representation}
\end{align}
Consider the case $N\gg n$. 
In this case, by using the saddle point evaluation, Eq.~(\ref{eqn:EmpiricalBayesLikelihood_ML-representation}) is reduced to
\begin{align*}
L_{\mrm{EB}}(H,\gamma)&\approx \frac{1}{n N} \ln P_{\mrm{prior}}(\hat{h}_{\mrm{ML}}\mid H) \nn
\aleq
+ \frac{1}{nN} \ln P_{\mrm{prior}}(\hat{\bm{J}}_{\mrm{ML}} \mid \gamma)+\mrm{constant}.
\end{align*}
In this case, the empirical Bayes' estimates $\{\hat{H},\hat{\gamma}\}$ thus converge to the maximum likelihood estimates of the hyperparameters in the prior distributions 
in which the maximum likelihood estimates of the parameters $\{\hat{h}_{\mrm{ML}},\hat{\bm{J}}_{\mrm{ML}}\}$ (i.e., the solution to the BML) are inserted.  
This indicates that the parameter estimations can be conducted independently of the hyperparameter estimation.
In this study, we do not concern ourselves with this trivial case.

\section{Statistical Mechanical Analysis}
\label{sec:StatisticalMechanicalAnalysis}

The empirical Bayes likelihood function in Eq.~(\ref{eqn:EmpiricalBayesLikelihood}) involves intractable multiple integrations.
In this section, we evaluate the empirical Bayes likelihood function using a statistical mechanical analysis. 
We consider the two types of the prior distribution of $\bm{J}$: one is the Gaussian prior in Eq.~(\ref{eqn:GaussPrior}),  
and the other is the Laplace prior in Eq.~(\ref{eqn:LaplacePrior}).

First, we evaluate the empirical Bayes likelihood function on the basis of the Gaussian prior in Secs.~\ref{sec:ReplicaMethod}--\ref{sec:algorithm}, 
after which we describe the evaluation based on the Laplace prior in Sec.~\ref{sec:Analysis_LaplacePrior}.

\subsection{Replica method}
\label{sec:ReplicaMethod}

The empirical Bayes likelihood function in Eq.~(\ref{eqn:EmpiricalBayesLikelihood}) can be represented as
\begin{align}
L_{\mrm{EB}}(H,\gamma)=\frac{1}{nN}\ln \lim_{x \to -1} \Psi_x(H,\gamma),
\label{eqn:EmpiricalBayesLikelihood_ReplicaMethod}
\end{align}
where
\begin{align}
&\Psi_x(H,\gamma)\nn
&:=\Big[ Z(h,\bm{J})^{x N} \exp N\Big(h \sum_{i=1}^n d_i + \sum_{i<j}J_{ij} d_{ij}\Big) \Big]_{h, \bm{J}},
\label{eqn:def_Psi_x}
\end{align}
and 
\begin{align*}
d_i := \frac{1}{N} \sum_{\mu = 1}^N \mrm{S}_i^{(\mu)},\quad  d_{ij} := \frac{1}{N} \sum_{\mu = 1}^N \mrm{S}_i^{(\mu)}\mrm{S}_j^{(\mu)}
\end{align*}
are the sample averages of the observed data points. 
We assume that $\tau_x := x N$ is a natural number, and therefore Eq.~(\ref{eqn:def_Psi_x}) can be expressed as 
\begin{align}
\Psi_x(H,\gamma)
&=\Big[ \sum_{\mcal{S}_x}\exp \Big\{h \sum_{i=1}^n\Big( \sum_{a = 1}^{\tau_x}S_i^{\{a\}}+ N d_i\Big) \nn
\aleq
+ \sum_{i<j}J_{ij}\Big(\sum_{a = 1}^{\tau_x}S_i^{\{a\}}S_j^{\{a\}} + N d_{ij}\Big)\Big\} \Big]_{h, \bm{J}},
\label{eqn:Psi_x_naturalnumberAssumption}
\end{align}
where $a ,b \in \{1,2,\ldots, \tau_x\}$ are replica indices, and $S_i^{\{a\}}$ is the Ising variable on site $i$ in the $a$th replica.
$\mcal{S}_x:= \{S_i^{\{a\}} \mid i = 1,2,\ldots, n;\, a = 1,2,\ldots, \tau_x\}$ is the set of all the Ising variables in the replicated system, and $\sum_{\mcal{S}_x}$ 
is the sum over all the possible configurations of $\mcal{S}_x$; i.e., $\sum_{\mcal{S}_x} := \prod_{i=1}^n\prod_{a=1}^{\tau_x} \sum_{S_i^{\{a\}} = \pm 1}$.
We evaluate $\Psi_x(H,\gamma)$ under the assumption that $\tau_x$ us a natural number, 
after which we take the limit of $x \to -1$ of the evaluation result to obtain the empirical Bayes likelihood function (this is the so-called \textit{replica trick}).

By employing the Gaussian prior in Eq.~(\ref{eqn:GaussPrior}), Eq.~(\ref{eqn:Psi_x_naturalnumberAssumption}) becomes 
\begin{align}
\Psi_x^{\mrm{Gauss}}(H,\gamma)
&=\exp\Big\{ n N H M + \frac{\gamma (n-1) N^2}{4}\Big(C_2 + \frac{x}{N}\Big)\nn
\aleq
 - F_x(H, \gamma)\Big\},
\label{eqn:Psi_x_Gauss}
\end{align}
where
\begin{align}
M:= \frac{1}{n}\sum_{i=1}^n d_i,\quad
C_k:= \frac{2}{n(n-1)}\sum_{i<j}d_{ij}^k,
\label{eqn:def_M&Ck}
\end{align}
and
\begin{align}
F_x(H, \gamma):=-\ln \sum_{\mcal{S}_x}\exp\big(-E_x(\mcal{S}_x;H,\gamma)\big)
\label{eqn:ReplicatedFreeEnergy}
\end{align}
is the replicated (Helmholtz) free energy~\cite{RCVM2010,YKT2012,RCVM2013,Yasuda2015}; 
here,  
\begin{align}
&E_x(\mcal{S}_x;H,\gamma)\nn
&:=-H \sum_{i=1}^n\sum_{a=1}^{\tau_x} S_i^{\{a\}}
- \frac{\gamma N}{n}\sum_{i<j}d_{ij} \sum_{a = 1}^{\tau_x}S_i^{\{a\}}S_j^{\{a\}} \nn
\aldef
- \frac{\gamma}{n}\sum_{i<j}\sum_{a < b}S_i^{\{a\}}S_j^{\{a\}}S_i^{\{b\}}S_j^{\{b\}}
\label{eqn:ReplicatedHamiltonian}
\end{align}
is the Hamiltonian of the replicated system, 
where $\sum_{a<b}$ is the sum over all the distinct pairs of replicas; i.e., $\sum_{a<b} = \sum_{a=1}^{\tau_x}\sum_{b = a+1}^{\tau_x}$. 

\subsection{Plefka expansion}
\label{sec:PlefkaExpansion}

Because the replicated free energy in Eq.~(\ref{eqn:ReplicatedFreeEnergy}) includes intractable multiple summations, 
an approximation is needed to proceed with our evaluation.
In this section, we approximate the replicated free energy using the Plefka expansion~\cite{Plefka1982}.
In brief, the Plefka expansion is the perturbative expansion in a Gibbs free energy that is a dual form of a corresponding Helmholtz free energy. 

The Gibbs free energy is obtained as 
\begin{align}
G_x(m,H,\gamma)
&=- n \tau_x H m + \extr_{\lambda}\Big\{\lambda n \tau_x m \nn
\aleq
-\ln \sum_{\mcal{S}_x}\exp\big( - E_x(\mcal{S}_x;\lambda,\gamma)\big)\Big\}.
\label{eqn:GibbsFreeEnergy}
\end{align} 
The derivation of this Gibbs free energy is described in Appendix \ref{sec:app:GibbsFreeEnergy}. 
It is noteworthy that this type of expression of the Gibbs free energy implies the replica-symmetric (RS) assumption. 
To take the replica-symmetry breaking (RSB) into account, explicit treatments of overlaps between different replicas are needed~\cite{YKT2012}.
By expanding $G_x(m,H,\gamma)$ around $\gamma = 0$, we obtain
\begin{align}
\frac{G_x(m,H,\gamma)}{nN} &=-x H m+ x e(m) + \phi_x^{(1)}(m) \gamma \nn
\aleq
+ \phi_x^{(2)}(m)\gamma^2 + O(\gamma^3),
\label{eqn:PlefkaExpansion}
\end{align}
where $e(m)$ is the negative mean-field entropy defined by
\begin{align}
e(m):=\frac{1+m}{2} \ln \frac{1+m}{2} +  \frac{1-m}{2} \ln \frac{1-m}{2},  
\label{eqn:MeanFieldEntropy}
\end{align}
and the coefficients, $\phi_x^{(1)}(m)$ and $\phi_x^{(2)}(m)$, are expressed as Eqs.~(\ref{eqn:PlefkaExpansion_1st}) and (\ref{eqn:PlefkaExpansion_2nd}), respectively. 
The detailed derivation of these coefficients is presented in Appendix \ref{sec:app:PlefkaExpansion}.

From Eqs.~(\ref{eqn:EmpiricalBayesLikelihood_ReplicaMethod}), (\ref{eqn:Psi_x_Gauss}), (\ref{eqn:PlefkaExpansion}), and (\ref{eqn:relation_F&G}),  
we obtain the empirical Bayes likelihood function as 
\begin{align}
L_{\mrm{EB}}(H,\gamma)&\approx HM  -\extr_{m}\Big[ Hm - e(m)+ \Phi(m)\gamma\nn
\aleq
+\phi_{-1}^{(2)}(m)\gamma^2\Big].
\label{eqn:EmpiricalBayesLikelihood_Gauss_Result}
\end{align}
where
\begin{align*}
\Phi(m):=\phi_{-1}^{(1)}(m) - \frac{(n-1)N}{4n}\Big(C_2 - \frac{1}{N}\Big).
\end{align*}
From Eqs.~(\ref{eqn:PlefkaExpansion_1st}) and (\ref{eqn:PlefkaExpansion_2nd}), $\Phi(m)$ and $\phi_{-1}^{(2)}(m)$ are
\begin{align}
\Phi(m) &= \frac{(n-1)NC_1}{2n}m^2- \frac{(n-1)N}{4n}\Big\{C_2  \nn
\aleq
+ \frac{N+1}{N}\Big(m^4 - \frac{1}{N+1}\Big)\Big\}
\label{eqn:Phi(m)}
\end{align}
and
\begin{widetext}
\begin{align}
\phi_{-1}^{(2)}(m)&=\frac{(n-1)^2  N^2 \Omega}{2n^2}m^2(1 - m^2)
+\frac{(n-1) N^2 C_2}{4n^2}(1-m^2)^2 - \frac{(n-1)N(N+1)C_1}{2n^2}m^2(1-m^2)^2\nn
\aleq
-\frac{(n-1) (N+1)}{4n^2}\big(n - N - 3\big)m^4(1-m^2)^2
-\frac{(n-1)(N+1)}{8n^2}(1 - m^4)^2,
\label{eqn:PlefkaExpansion_2nd_x=-1}
\end{align}
\end{widetext}
respectively. The coefficient $\Omega$ appearing in the above equation is defined by 
\begin{align}
\Omega:=\frac{1}{n}\sum_{i=1}^n \omega_i^2,
\label{eqn:def_Omega}
\end{align}
where
\begin{align}
\omega_i &:=\frac{1}{n-1}\sum_{j\in \partial(i)}d_{ij} - C_1;
\label{eqn:def_omega_i}
\end{align}
here, $\partial(i):= \{1,2,\ldots,n\} \setminus \{i\}$.

\subsection{Inference algorithm}
\label{sec:algorithm}

As mentioned in Sec.~\ref{sec:Framework_EB}, the empirical Bayes inference is achieved by maximizing $L_{\mrm{EB}}(H,\gamma)$ with respect to $H$ and $\gamma$ (cf. Eq.~(\ref{eqn:Maximizing_EBL})). 
From the extremum condition of Eq.~(\ref{eqn:EmpiricalBayesLikelihood_Gauss_Result}) with respect to $H$, we obtain 
\begin{align}
\hat{m} = M,
\label{eqn:Extr_H}
\end{align}
where $\hat{m}$ is the value of $m$ that satisfies the extremum condition in Eq.~(\ref{eqn:EmpiricalBayesLikelihood_Gauss_Result}). 
From the extremum condition of Eq.~(\ref{eqn:EmpiricalBayesLikelihood_Gauss_Result}) with respect to $m$ and Eq.~(\ref{eqn:Extr_H}), 
we obtain
\begin{align}
\hat{H} =\tanh^{-1}M -\Big(\frac{\partial \phi_{-1}^{(1)}(m)}{\partial m}\gamma  
+\frac{\partial \phi_{-1}^{(2)}(m)}{\partial m}\gamma^2\Big)
\Big|_{m = M}.
\label{eqn:determine_H-hat}
\end{align}
From Eqs.~(\ref{eqn:EmpiricalBayesLikelihood_Gauss_Result}) and (\ref{eqn:Extr_H}), the optimal value of $\gamma$ is obtained by
\begin{align}
\hat{\gamma}&=\argmax_{\gamma}\big[-\Phi(M)\gamma -\phi_{-1}^{(2)}(M)\gamma^2\big].
\label{eqn:determine_gamma-hat}
\end{align}
From Eq.~(\ref{eqn:determine_gamma-hat}), $\hat{\gamma}$ is immediately obtained as follows: 
(i) when $\phi_{-1}^{(2)}(M) > 0$ and $\Phi(M) \geq  0$ or when $\phi_{-1}^{(2)}(M) = 0$ and $\Phi(M) > 0$, $\hat{\gamma} = 0$, 
(ii) when $\phi_{-1}^{(2)}(M) > 0$ and $\Phi(M) < 0$, $\hat{\gamma} = - \Phi(M) / (2 \phi_{-1}^{(2)}(M))$, 
and (iii) $\hat{\gamma} \to \infty$ elsewhere.
Here, we ignore the case $\phi_{-1}^{(2)}(M) = \Phi(M) = 0$, because it hardly occurs in realistic settings.  
By using Eqs.~(\ref{eqn:determine_H-hat}) and (\ref{eqn:determine_gamma-hat}), 
we can obtain the solution to the empirical Bayes inference without any iterative processes. 
The pseudocode of the proposed procedure is shown in Algorithm \ref{alg:EmpiricalBayes}. 
\begin{algorithm}[H]
\caption{Proposed Inference Algorithm}
\label{alg:EmpiricalBayes}
\begin{algorithmic}[1]
\State \textbf{Input} Observed data set: $\mcal{D}:=\{\mbf{S}^{(\mu)} \in \{-1,+1\}^n \mid \mu = 1,2,\ldots, N\}$.
\State Compute $M$, $\Omega$, $C_1$, and $C_2$ using the data set according to Eqs.~(\ref{eqn:def_M&Ck}) and (\ref{eqn:def_Omega}).
\State Determine $\hat{\gamma}$ using Eq.~(\ref{eqn:determine_gamma-hat}):
\begin{align*}
\hat{\gamma}=
\begin{cases}
0 & \mbox{case (i)}\\
- \Phi(M) / (2 \phi_{-1}^{(2)}(M)) & \mbox{case (ii)}\\
\infty & \mrm{elsewhere}
\end{cases}
,
\end{align*}
where case (i): $\phi_{-1}^{(2)}(M) > 0, \>\Phi(M) \geq  0$ or $\phi_{-1}^{(2)}(M) = 0, \> \Phi(M) > 0$ 
and case (ii): $\phi_{-1}^{(2)}(M) > 0, \> \Phi(M) < 0$.
\State Using $\hat{\gamma}$, determine $\hat{H}$ using Eq.~(\ref{eqn:determine_H-hat}).
\State \textbf{Output} $\hat{\gamma}$ and $\hat{H}$.
\end{algorithmic}
\end{algorithm}

In the proposed method, the value of $\hat{H}$ does not affect the determination of $\hat{\gamma}$. 
Many mean-field-based methods for BML (e.g., listed in Sec.~\ref{sec:intro}) have similar procedures, 
in which $\hat{\bm{J}}_{\mrm{ML}}$ are determined separately from $\hat{h}_{\mrm{ML}}$.  
This is seen as one of the common properties of the mean-field-based methods for BML including the current empirical Bayes problem.

\subsection{Evaluation based on Laplace prior}
\label{sec:Analysis_LaplacePrior}

The above evaluation was for the Gaussian prior in Eq.~(\ref{eqn:GaussPrior}). 
Here, we explain the evaluation for the Laplace prior in Eq.~(\ref{eqn:LaplacePrior}).
By employing the Laplace prior in Eq.~(\ref{eqn:LaplacePrior}), Eq.~(\ref{eqn:Psi_x_naturalnumberAssumption}) becomes 
\begin{align}
&\Psi_x^{\mrm{Laplace}}(H,\gamma)\nn
&=\xi^{n(n-1)}e^{nNHM}\sum_{\mcal{S}_x}\exp\Big[ H\sum_{i=1}^n\sum_{a = 1}^{\tau_x}S_i^{\{a\}} \nn
\aleq
- \sum_{i<j}\ln \Big\{ \xi^2 -\Big(\sum_{a=1}^{\tau_x}S_i^{\{a\}}S_j^{\{a\}}+ N d_{ij}\Big)^2\Big\}\Big],
\label{eqn:Psi_x_Laplace}
\end{align}
where $\xi:=\sqrt{2n/\gamma}$. Here, we assume
\begin{align}
\xi > \max_{i<j}\Big(\sum_{a=1}^{\tau_x}S_i^{\{a\}}S_j^{\{a\}}+ N d_{ij}\Big).
\label{eqn:assumption_Laplace}
\end{align}
By using the perturbative approximation, 
\begin{align*}
&\ln \Big\{ \xi^2 -\Big(\sum_{a=1}^{\tau_x}S_i^{\{a\}}S_j^{\{a\}}+ N d_{ij}\Big)^2\Big\}\nn
&=  \ln \xi^2 + \ln \Big\{1 -\xi^{-2}\Big(\sum_{a=1}^{\tau_x}S_i^{\{a\}}S_j^{\{a\}}+ N d_{ij}\Big)^2\Big\}\nn
&\approx 
\ln \xi^2 -\xi^{-2}\Big(\sum_{a=1}^{\tau_x}S_i^{\{a\}}S_j^{\{a\}}+ N d_{ij}\Big)^2,
\end{align*}
we obtain the approximation of Eq.~(\ref{eqn:Psi_x_Laplace}) as
\begin{align*}
\Psi_x^{\mrm{Laplace}}(H,\gamma)
&\approx e^{nNHM}\sum_{\mcal{S}_x}\exp\Big[ H\sum_{i=1}^n\sum_{a = 1}^{\tau_x}S_i^{\{a\}} \nn
\aleq
+ \xi^2\sum_{i<j}\Big(\sum_{a=1}^{\tau_x}S_i^{\{a\}}S_j^{\{a\}}+ N d_{ij}\Big)^2\Big],
\end{align*}
The right-hand side of this equation coincides with $\Psi_x^{\mrm{Gauss}}(H,\gamma)$ in Eq.~(\ref{eqn:Psi_x_Gauss}). 
This means that the empirical Bayes inference based on the Laplace prior in Eq.~(\ref{eqn:LaplacePrior}) is 
(approximately) equivalent to that based on the Gaussian prior in Eq.~(\ref{eqn:GaussPrior})  
(i.e., $\Psi_x^{\mrm{Laplace}}(H,\gamma) \approx \Psi_x^{\mrm{Gauss}}(H,\gamma)$) when the assumption of Eq.~(\ref{eqn:assumption_Laplace}) is justified.
Thus, we can also use the algorithm presented in Sec.~\ref{sec:algorithm} for the case of the Laplace prior.

\section{Numerical Experiments}
\label{sec:experiment}

In this section, we describe the results of our numerical experiments. 
In these experiments, the observed dataset $\mcal{D}$ are generated from 
the generative Boltzmann machine, which has the same form as Eq.~(\ref{eqn:BoltzmannMachine}), by using annealed importance sampling (AIS)~\cite{Neal2001}. 
In AIS, we controlled the annealing schedule using a series of inverse temperature $0 = \beta_0 < \beta_1 < \cdots < \beta_T = 1$, where $\beta_{t + 1} = \beta_t + 0.03$.
The parameters of the generative Boltzmann machine are drawn from the prior distributions in Eqs.~(\ref{eqn:prior_J}) and (\ref{eqn:prior_H}). 
That is, we consider the model-matched case (i.e., the generative and learning models are identical). 

In the following, we use the notations $\alpha := N / n$ and $J := \sqrt{\gamma}$. 
The standard deviations of the Gaussian prior in Eq.~(\ref{eqn:GaussPrior}) and of the Laplace prior in Eq.~(\ref{eqn:LaplacePrior}) are then $J / \sqrt{n}$. 
We express the hyperparameters for the generative Boltzmann machine by $H_{\mrm{true}}$ and $J_{\mrm{true}}$. 

\subsection{Gaussian prior case}

Here, we consider the case in which the prior distribution of $\bm{J}$ is the Gaussian prior in Eq.~(\ref{eqn:GaussPrior}). 
In this case, the Boltzmann machine corresponds to the Sherrington-Kirkpatrick (SK) model~\cite{SK1975}, 
and therefore it shows the spin-glass transition at $J = 1$ when $h = 0$ (i.e., when $H = 0$).

\begin{figure}[tb]
\centering
\includegraphics[height=4.4cm]{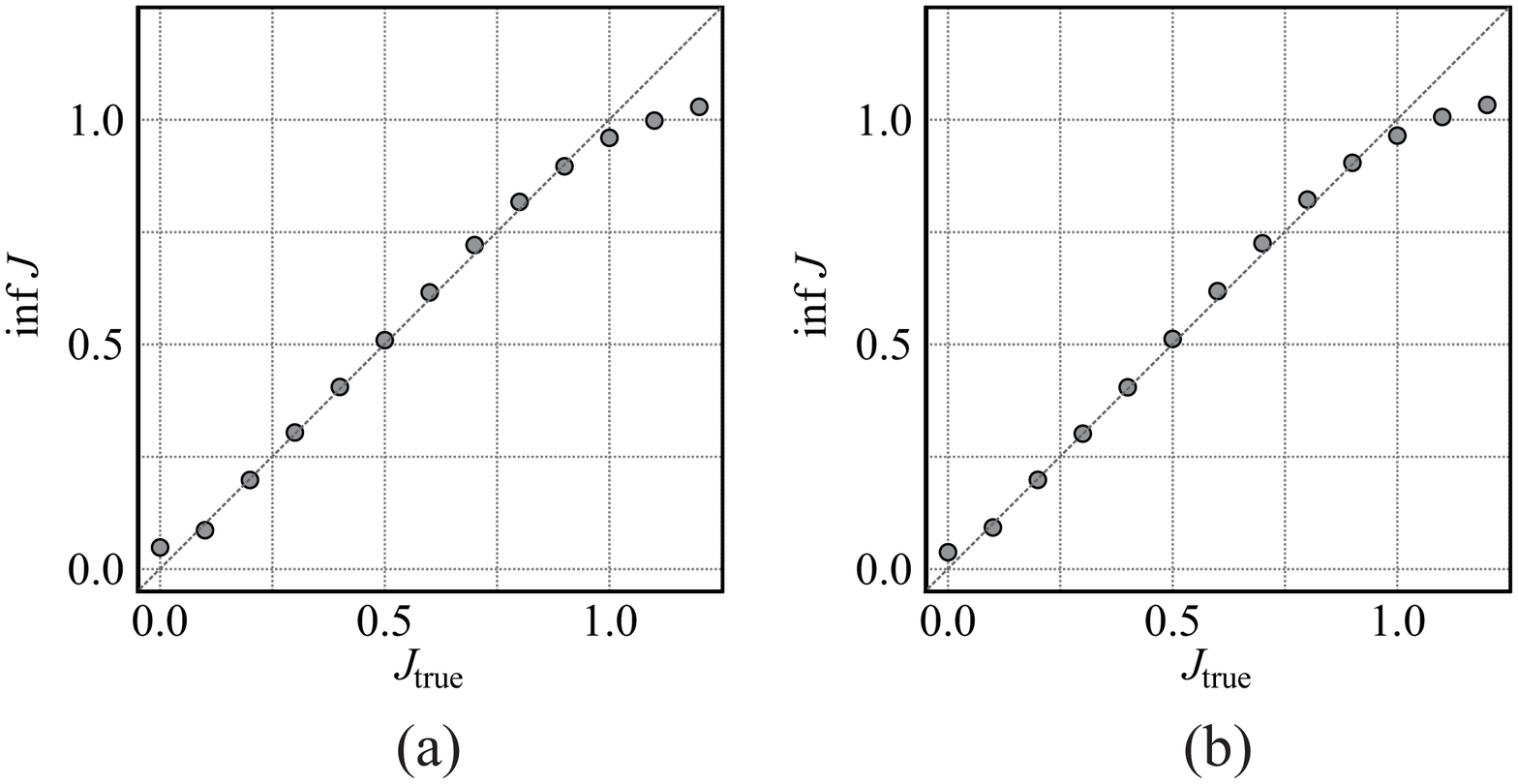}
\caption{Scatter plots of $J_{\mrm{true}}$ (horizontal axis) versus $\hat{J}$ (vertical axis) when $H_{\mrm{true}} = 0$ and $\alpha = 0.4$: (a) $n = 300$ and (b) $n = 500$.
Plots are the average values over 300 experiments.}
\label{fig:J_Gauss_alpha0.4}
\end{figure}

\begin{figure}[tb]
\centering
\includegraphics[height=4.4cm]{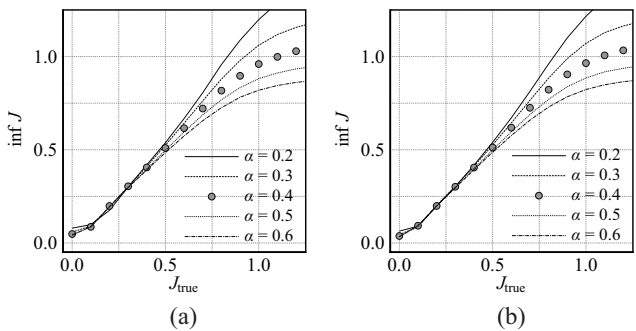}
\caption{Scatter plots of $J_{\mrm{true}}$ (horizontal axis) versus $\hat{J}$ (vertical axis) for various $\alpha = N / n$ when $H_{\mrm{true}} = 0$: (a) $n = 300$ and (b) $n = 500$.
Plots are the average values over 300 experiments.}
\label{fig:J_Gauss}
\end{figure}

\begin{figure}[tb]
\centering
\includegraphics[height=4.4cm]{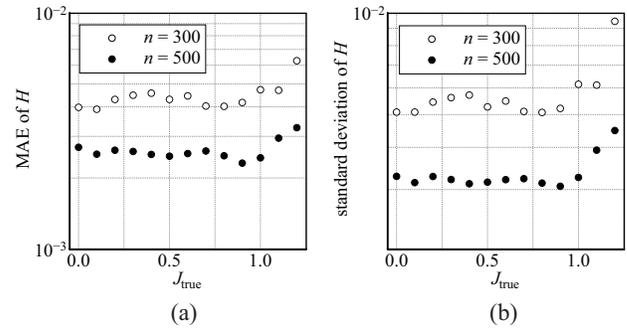}
\caption{Results of estimation of $\hat{H}$ against $J_{\mrm{true}}$ when $H_{\mrm{true}} = 0$ and $\alpha = 0.4$: (a) the mean absolute error and (b) standard deviation.
Plots are the average values over 300 experiments.}
\label{fig:H_Gauss_alpha0.4}
\end{figure}

First, we consider the case $H_{\mrm{true}} = 0$.
We show the scatter plots for the estimation of $\hat{J}$ for various $J_{\mrm{true}}$ when $H_{\mrm{true}} = 0$ and $\alpha = 0.4$ in Fig.~\ref{fig:J_Gauss_alpha0.4}.
The detailed values of the plots for some $J_{\mrm{true}}$ values are shown in Tab.~\ref{tab:J_Gauss_alpha0.4}.
\begin{table*}[]
\caption{Detailed values (the averages and standard deviations) of some plots in Fig.~\ref{fig:J_Gauss_alpha0.4} (when $H_{\mrm{true}} = 0$ and $\alpha = 0.4$).}
\begin{tabular}{cc|c|c|c|c|c|c|c|}
\cline{3-9}
 &  & \multicolumn{7}{c|}{$J_{\mrm{true}}$} \\ \cline{3-9} 
 &  & 0 & 0.2 & 0.4 & 0.6 & 0.8 & 1 & 1.2 \\ \hline
\multicolumn{1}{|c|}{\multirow{2}{*}{$\hat{J}$}} & $n = 300$ & $0.048 \pm 0.06$ & $0.20 \pm 0.04$ & $0.41 \pm 0.02$ & $0.62 \pm 0.02$ & $0.82 \pm 0.02$  & $0.96 \pm 0.02$ & $1.03 \pm 0.02$\\ \cline{2-9} 
\multicolumn{1}{|c|}{} & $n = 500$ & $0.038 \pm 0.05$ & $0.20 \pm 0.03$ & $0.40 \pm 0.01$ & $0.62 \pm 0.01$ & $0.82 \pm 0.01$  & $0.96 \pm 0.01$ & $1.03 \pm 0.01$ \\ \hline
\end{tabular}
\label{tab:J_Gauss_alpha0.4}
\end{table*}
When $J_{\mrm{true}} < 1$, our estimates of $\hat{J}$ are in good agreement with $J_{\mrm{true}}$. 
This implies that the validity of our perturbative approximation is lost in the spin-glass phase, as is often the case with many mean-field approximations.
Fig.~\ref{fig:J_Gauss} shows the scatter plots for various $\alpha$.
A smaller $\alpha$ causes $\hat{J}$ to be overestimated and a larger $\alpha$ causes it to be underestimated. 
At least in our experiments, the optimal value of $\alpha$ seems to be $\alpha_{\mrm{opt}} \approx 0.4$ when $H_{\mrm{true}} = 0$. 
Our method can estimate $\hat{H}$ together with $\hat{J}$. 
The results for the estimation of $\hat{H}$ when $H_{\mrm{true}} = 0$ and $\alpha = 0.4$ are shown in Fig.~\ref{fig:H_Gauss_alpha0.4}.
Figs.~\ref{fig:H_Gauss_alpha0.4}(a) and (b) show the average of $|H_{\mrm{true}} - \hat{H}|$ (i.e., the mean absolute error (MAE))
and the standard deviation of $\hat{H}$ over 300 experiments, respectively. 
The MAE and standard deviation increase in the region $J_{\mrm{true}} > 1$.

\begin{figure}[tb]
\centering
\includegraphics[height=4.4cm]{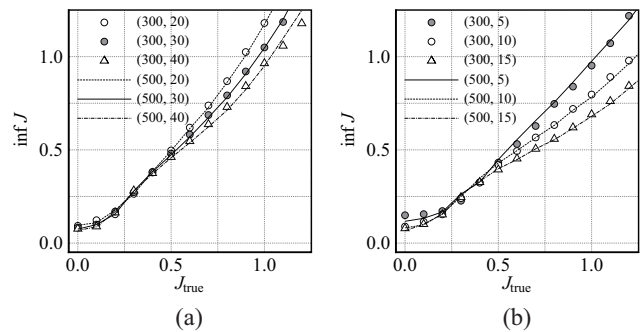}
\caption{Scatter plots of $J_{\mrm{true}}$ (horizontal axis) versus $\hat{J}$ (vertical axis) for various $\alpha = N / n$ for $n = 300$ and $500$: (a) $H_{\mrm{true}} = 0.2$ and (b) $H_{\mrm{true}} = 0.4$.
Plots are the average values over 300 experiments. The notation in the legend means $(n, N)$.}
\label{fig:J_Gauss_H0.2&0.4}
\end{figure}

\begin{figure}[tb]
\centering
\includegraphics[height=4.4cm]{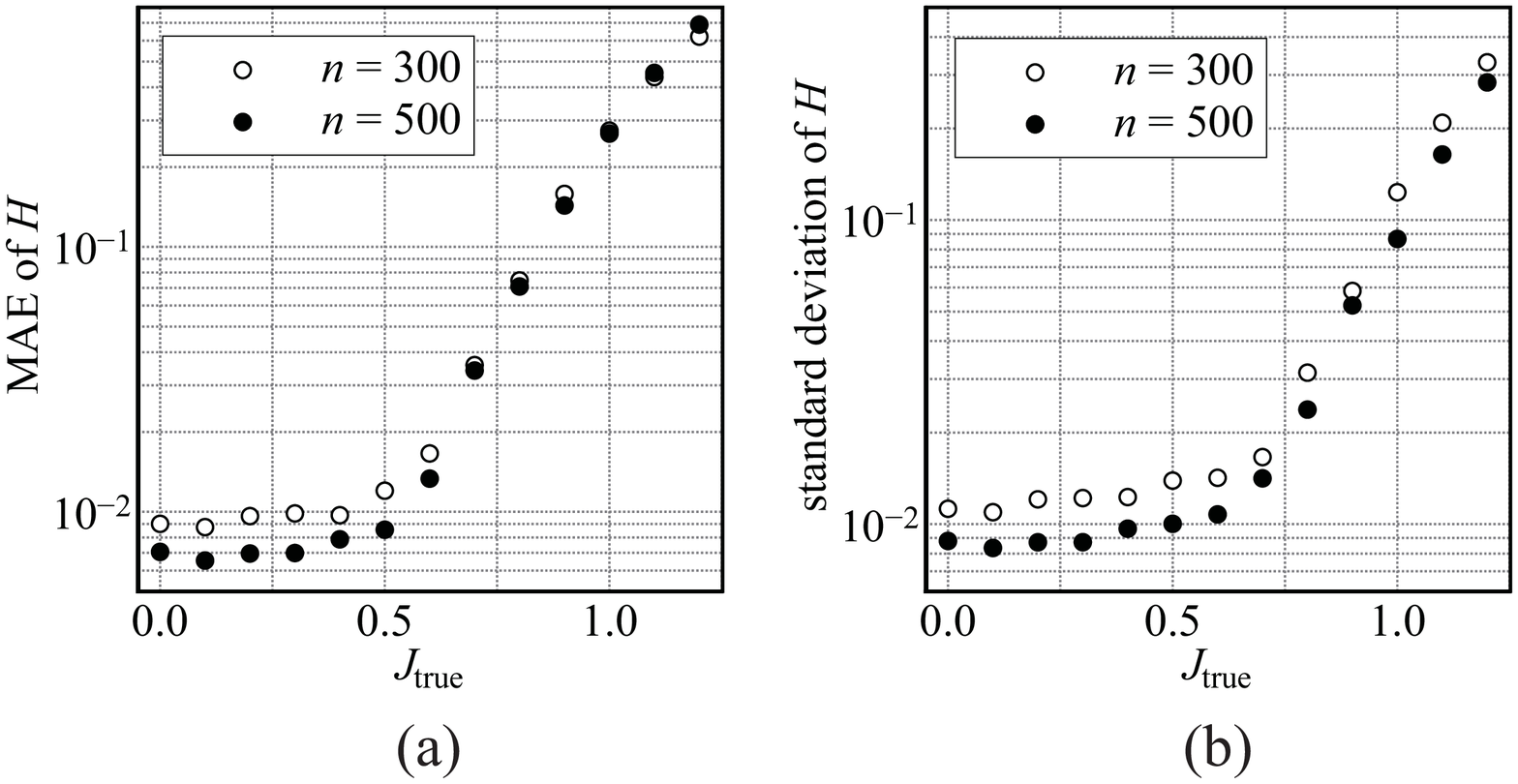}
\caption{Results of estimation of $\hat{H}$ against $J_{\mrm{true}}$ when $H_{\mrm{true}} = 0.2$ and $\alpha = 30/ n$: (a) the mean absolute error and (b) standard deviation.
Plots are the average values over 300 experiments.}
\label{fig:H_Gauss_H0.2}
\end{figure}

\begin{figure}[tb]
\centering
\includegraphics[height=4.4cm]{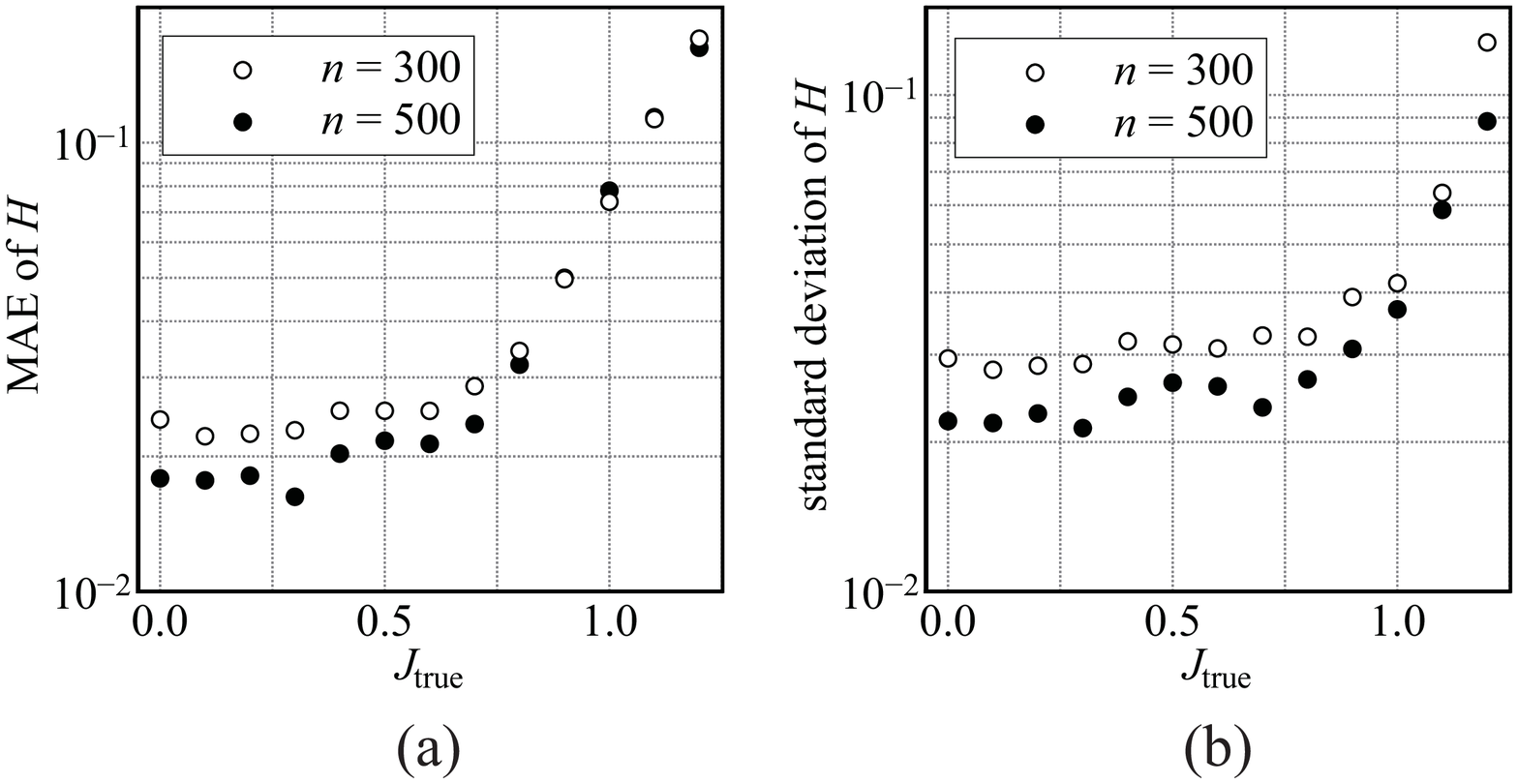}
\caption{Results of estimation of $\hat{H}$ against $J_{\mrm{true}}$ when $H_{\mrm{true}} = 0.4$ and $\alpha = 5 / n$: (a) the mean absolute error and (b) standard deviation.
Plots are the average values over 300 experiments.}
\label{fig:H_Gauss_H0.4}
\end{figure}

Next, we consider the cases $H_{\mrm{true}} > 0$. 
The scatter plots for the estimation of $\hat{J}$ for various $J_{\mrm{true}}$ values when $H_{\mrm{true}} = 0.2$ and $H_{\mrm{true}} =0.4$ are shown in Fig.~\ref{fig:J_Gauss_H0.2&0.4}.
The appropriate values of $\alpha$ when $H_{\mrm{true}} = 0.2$ and $H_{\mrm{true}} = 0.4$ ``approximately'' seem to be $\alpha_{\mrm{opt}} \approx 30/n$ and $\alpha_{\mrm{opt}} \approx 5/n$, respectively.
The detailed values of these plots for some $J_{\mrm{true}}$ values are shown in Tabs.~\ref{tab:J_Gauss_H0.2_N30} and \ref{tab:J_Gauss_H0.4_N5}.
The results for the estimation of $\hat{H}$ when $H_{\mrm{true}} = 0.2$ and $\alpha = 30/n$ and when $H_{\mrm{true}} = 0.4$ and $\alpha = 5/n$ 
are shown in Figs.~\ref{fig:H_Gauss_H0.2} and \ref{fig:H_Gauss_H0.4}, respectively.
The increases in the MAE and standard deviations occur earlier than for the case in Fig.~\ref{fig:H_Gauss_alpha0.4}.

\begin{table*}[]
\caption{Detailed values (the averages and standard deviations) of some plots in Fig.~\ref{fig:J_Gauss_H0.2&0.4} (a) (when $H_{\mrm{true}} = 0.2$ and $\alpha = 30/n$).}
\begin{tabular}{cc|c|c|c|c|c|c|c|}
\cline{3-9}
 &  & \multicolumn{7}{c|}{$J_{\mrm{true}}$} \\ \cline{3-9} 
 &  & 0 & 0.2 & 0.4 & 0.6 & 0.8 & 1 & 1.2 \\ \hline
\multicolumn{1}{|c|}{\multirow{2}{*}{$\hat{J}$}} & $n = 300$ & $0.083 \pm 0.10$ & $0.17 \pm 0.12$ & $0.38 \pm 0.07$ & $0.58 \pm 0.05$ & $0.79 \pm 0.06$  & $1.05 \pm 0.12$ & $1.35 \pm 0.16$\\ \cline{2-9} 
\multicolumn{1}{|c|}{} & $n = 500$ & $0.075 \pm 0.09$ & $0.16 \pm 0.11$ & $0.38 \pm 0.06$ & $0.57 \pm 0.04$ & $0.78 \pm 0.06$  & $1.05 \pm 0.10$ & $1.39 \pm 0.16$ \\ \hline
\end{tabular}
\label{tab:J_Gauss_H0.2_N30}
\end{table*}
\begin{table*}[]
\caption{Detailed values (the averages and standard deviations) of some plots in Fig.~\ref{fig:J_Gauss_H0.2&0.4} (b) (when $H_{\mrm{true}} = 0.4$ and $\alpha = 5/n$).}
\begin{tabular}{cc|c|c|c|c|c|c|c|}
\cline{3-9}
 &  & \multicolumn{7}{c|}{$J_{\mrm{true}}$} \\ \cline{3-9} 
 &  & 0 & 0.2 & 0.4 & 0.6 & 0.8 & 1 & 1.2 \\ \hline
\multicolumn{1}{|c|}{\multirow{2}{*}{$\hat{J}$}} & $n = 300$ & $0.15 \pm 0.17$ & $0.17 \pm 0.17$ & $0.33 \pm 0.19$ & $0.53 \pm 0.14$ & $0.75 \pm 0.12$  & $0.95 \pm 0.14$ & $1.22 \pm 0.20$\\ \cline{2-9} 
\multicolumn{1}{|c|}{} & $n = 500$ & $0.12 \pm 0.15$ & $0.17 \pm 0.17$ & $0.33 \pm 0.17$ & $0.55 \pm 0.12$ & $0.76 \pm 0.10$  & $0.98 \pm 0.11$ & $1.20 \pm 0.16$ \\ \hline
\end{tabular}
\label{tab:J_Gauss_H0.4_N5}
\end{table*}

One of the largest qualitative differences between the cases $H_{\mrm{true}} = 0$ and $H_{\mrm{true}} > 0$ is the scale of $\alpha$.
In the case $H_{\mrm{true}} = 0$, the optimal $\alpha$ was scaled by $O(1)$ with respect to $n$ (i.e., $N = O(n)$). 
Meanwhile, in the case $H_{\mrm{true}} > 0$, the optimal $\alpha$ is scaled by $O(1/n)$ with respect to $n$ (i.e., $N = O(1)$). 
This change of scale can be understood from a scale evaluation for the terms in the empirical Bayes likelihood function in Eq.~(\ref{eqn:EmpiricalBayesLikelihood_Gauss_Result}).
The detailed reasoning is given in Appendix~\ref{sec:app:OrdarEvaluation}.

\subsection{Laplace prior case}

\begin{figure}[tb]
\centering
\includegraphics[height=4.4cm]{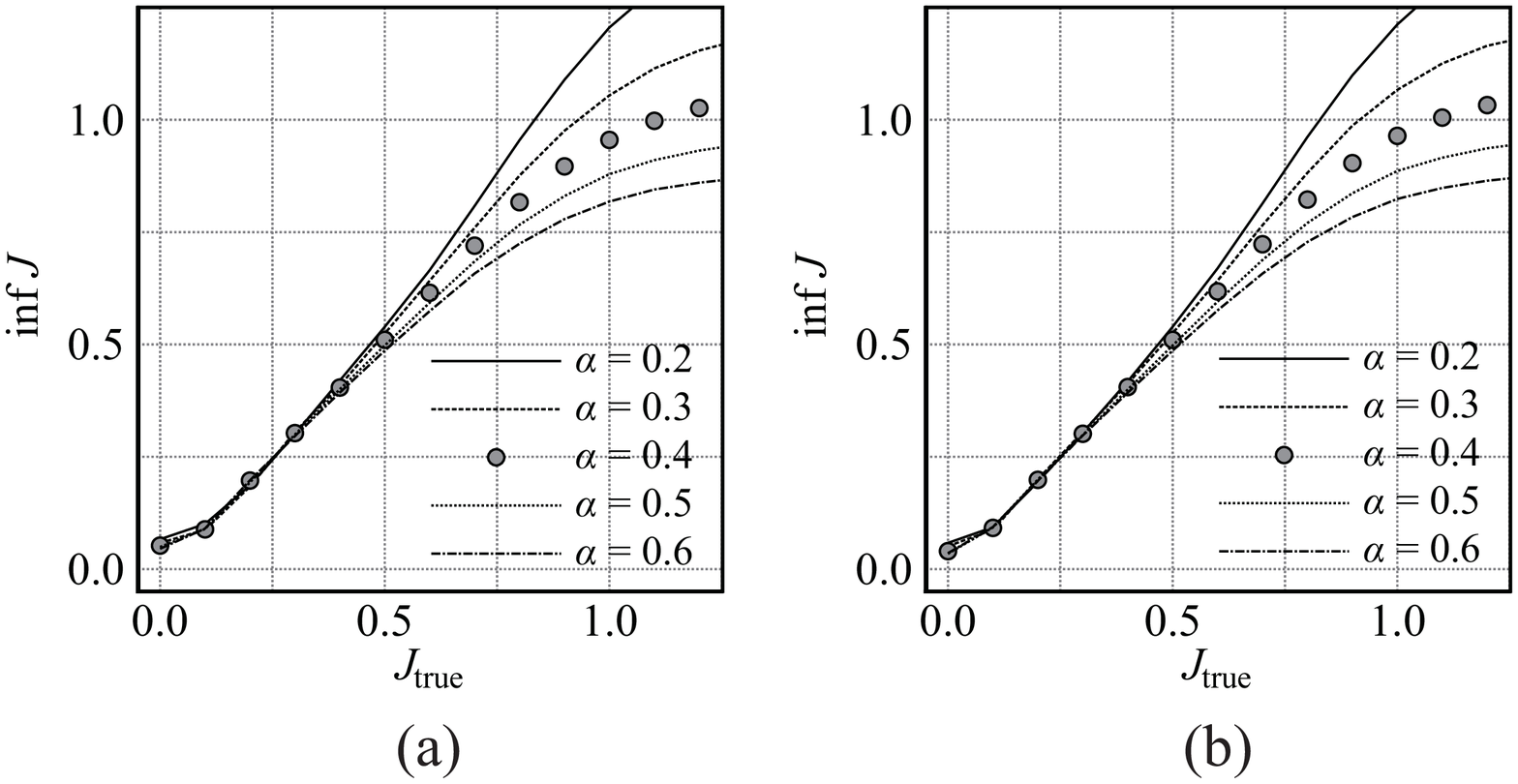}
\caption{Scatter plots of $J_{\mrm{true}}$ (horizontal axis) versus $\hat{J}$ (vertical axis) for various $\alpha = N/n$, when $H_{\mrm{true}} = 0$, 
in the case of the Laplace prior: (a) $n = 300$ and (b) $n = 500$.
Plots are the average values over 300 experiments.}
\label{fig:J_Laplace}
\end{figure}

Here, we consider the case in which the prior distribution of $\bm{J}$ is the Laplace prior in Eq.~(\ref{eqn:LaplacePrior}). 
The scatter plots for the estimation of $\hat{J}$ for various $J_{\mrm{true}}$ values when $H_{\mrm{true}} = 0$ are shown in Fig.~\ref{fig:J_Laplace}.
The plots shown in Fig.~\ref{fig:J_Laplace} almost completely overlap with those in Fig.~\ref{fig:J_Gauss}. 
Furthermore, all the numerical results in the case $H_{\mrm{true}} > 0$ also almost completely overlap with the corresponding results obtained in the above Gaussian prior case, 
and therefore we do not show those results. 

\section{Summary and Discussions}
\label{sec:summary}

In this study, we proposed a hyperparameters inference algorithm 
by analyzing the empirical Bayes likelihood function in Eq.~(\ref{eqn:EmpiricalBayesLikelihood}) 
using the replica method and the Plefka expansion. 
The validity of our method was examined in numerical experiments for the Gaussian and Laplace priors, 
which demonstrated the existence of an appropriate scale in the size of the dataset that can accurately recover the values of the hyperparameters.   

However, some problems remain. 
The first one is the scale of $N$. 
In our experiments, we found that an appropriate $N$ is scaled by $O(n)$ when $H_{\mrm{true}} = 0$ or by $O(1)$ when $H_{\mrm{true}} \neq 0$.
However, such scales seem to be unnatural, 
because they should not appear in the original framework of the empirical Bayes method. 
As discussed in Sec.~\ref{sec:Framework_EB}, when $N\gg n$, maximizing the empirical Bayes likelihood function 
is reduced to the maximum likelihood estimation of the prior distributions for the solution to BML. 
This must lead to the correct $\hat{\gamma}$ and $\hat{H}$, because the solution to BML is perfect when $N \to \infty$. 
Therefore, such unnatural scales appear due to our approximation, which is also supported by a scale analysis given in Appendix~\ref{sec:app:OrdarEvaluation}. 
An improvement of the approximation (e.g., by evaluating the leading terms in the Plefka expansion or using some other approximations) 
might reduce these unnatural behaviors. 

The second problem is the optimal setting $\alpha = N/n$. 
Empirically, we found that $\alpha_{\mrm{opt}} \approx 0.4$ when $H_{\mrm{true}} = 0$ 
and that it decreases as $H_{\mrm{true}}$ increases (e.g., $\alpha_{\mrm{opt}} \approx 30/n$ when $H_{\mrm{true}} = 0.2$ 
and $\alpha_{\mrm{opt}} \approx 5/n$ when $H_{\mrm{true}} = 0.4$). 
As can be seen in the results of our experiments, the solution to our method is robust for the choice of $\alpha$ when $J_{\mrm{true}}$ is small ($J_{\mrm{true}} < J_c$) 
and is sensitive to it when $J_{\mrm{true}}$ is large ($J_{\mrm{true}} > J_c$), where $J_c \approx 0.4$. 
The estimation of $\alpha_{\mrm{opt}}$ is very important for our method, 
and it will make our method more practical. 
This problem would be strongly related to the first problem. 

The third problem is the degradation of the estimation accuracy in the spin-glass phase. 
In our experiments, the estimation accuracies of $\hat{\gamma}$ and $\hat{H}$ were obviously degraded in the spin-glass phase.
This means that our Plefka expansion based on the RS assumption loses its validity in the spin-glass phase. 
In Ref.~\cite{YKT2012}, a Plefka expansion for the one-step RSB was proposed. 
Employing this expansion instead of the current expansion could reduce the degradation in the spin-glass phase.
These three problems should be addressed in our future studies.

In this study, we used fully-connected Boltzmann machines whose variables are all visible. 
We are also interested in an extension of our method to other types of Boltzmann machines 
such as Boltzmann machines having specific structures or hidden variables. 
Furthermore, we considered the model-matched case (i.e., the case in which the generative mode and learning model are the same model) 
in the current study, but model-mismatched cases are more practical and important.  

\appendix

\section{Gibbs Free Energy}
\label{sec:app:GibbsFreeEnergy}

In this appendix, we derive the Gibbs free energy for the replicated (Helmholtz) free energy in Eq.~(\ref{eqn:ReplicatedFreeEnergy}). 

The replicated free energy is obtained by minimizing the variational free energy, defined by
\begin{align}
f[Q]:=\sum_{\mcal{S}_x}E_x(\mcal{S};H,\gamma)Q(\mcal{S}_x) + \sum_{\mcal{S}_x}Q(\mcal{S}_x)\ln Q(\mcal{S}_x),
\label{eqn:VariationalFreeEnergy}
\end{align}
under the normalization constraint, i.e., $\sum_{\mcal{S}_x}Q(\mcal{S}_x) = 1$, 
where $Q(\mcal{S}_x)$ is a test distribution over $\mcal{S}_x$, and $E_x(\mcal{S}_x;H,\gamma)$ is the Hamiltonian for the replicated system defined in Eq.~(\ref{eqn:ReplicatedHamiltonian}).

The Gibbs free energy is obtained by adding new constraints to the minimization of $f[Q]$. 
Here, we add the relation $(n\tau_x)^{-1}\sum_{i=1}^n\sum_{a = 1}^{\tau_x}\sum_{\mcal{S}_x}S_i^{\{a\}}Q(\mcal{S}_x) = m$ as the constraint. 
By using Lagrange multipliers, the Gibbs free energy is obtained as
\begin{align}
G_x(m,H,\gamma)&:=\extr_{Q,\lambda, r} \Big\{ f[Q] - r \Big(\sum_{\mcal{S}_x}Q(\mcal{S}_x) - 1\Big)\nn
&-\lambda \Big(\sum_{i=1}^n\sum_{a = 1}^{\tau_x}\sum_{\mcal{S}_x}S_i^{\{a\}}Q(\mcal{S}_x) - n \tau_x m\Big)\Big\},
\label{eqn:def_GibbsFreeEnergy}
\end{align}
where ``$\extr$'' denotes the extremum with respect to the assigned parameters.
By performing the extremum operation with respect to $Q(\mcal{S})$ and $r$ in Eq.~(\ref{eqn:def_GibbsFreeEnergy}), we obtain
\begin{align}
&G_x(m,H,\gamma) \nn
&= \extr_{\lambda}\Big\{\lambda n \tau_x m 
-\ln \sum_{\mcal{S}_x}\exp\big( - E_x(\mcal{S}_x;H+\lambda,\gamma)\big)\Big\}.
\label{eqn:def_GibbsFreeEnergy_trans}
\end{align} 
The replicated free energy in Eq.~(\ref{eqn:ReplicatedFreeEnergy}) coincides with the extremum of this Gibbs free energy with respect to $m$; i.e., 
\begin{align}
F_x(H, \gamma) = \extr_{m}G_x(m,H,\gamma).
\label{eqn:relation_F&G}
\end{align}
By performing the shift $H + \lambda \to \lambda$ in Eq.~(\ref{eqn:def_GibbsFreeEnergy_trans}), we obtain Eq.~(\ref{eqn:GibbsFreeEnergy}). 

\section{Derivation of Coefficients of Plefka Expansion}
\label{sec:app:PlefkaExpansion}

The Plefka expansion considered in this study can be obtained by expanding the Gibbs free energy in Eq.~(\ref{eqn:GibbsFreeEnergy}) around $\gamma = 0$.

When $\gamma = 0$, we have
\begin{align}
G_x(m,H,0) &=- n \tau_x H m+ n \tau_x\extr_{\lambda}\big(\lambda m - \ln 2\cosh \lambda\big) \nn
&=- n \tau_x H m + n \tau_x e(m),
\label{eqn:PlefkaExpansion_0th}
\end{align}
where $e(m)$ is defined in Eq.~(\ref{eqn:MeanFieldEntropy}).

For the derivations of the coefficients $\phi_x^{(1)}(m)$ and $\phi_x^{(2)}(m)$, 
we decompose $E_x(\mcal{S}_x;H,\lambda)$ in Eq.~(\ref{eqn:GibbsFreeEnergy}) into two parts:
\begin{align*}
E_x(\mcal{S}_x;\lambda,\gamma)=-\lambda \sum_{i=1}^n\sum_{a=1}^{\tau_x} S_i^{\{a\}}
+ \gamma E_x^{\mrm{int}}(\mcal{S}_x),
\end{align*}
where
\begin{align*}
E_x^{\mrm{int}}(\mcal{S}_x)&:=- \frac{N}{n}\sum_{i<j}d_{ij} \sum_{a = 1}^{\tau_x}S_i^{\{a\}}S_j^{\{a\}} \nn
\aldef
- \frac{1}{n}\sum_{i<j}\sum_{a < b}S_i^{\{a\}}S_j^{\{a\}}S_i^{\{b\}}S_j^{\{b\}}.
\end{align*}
Coefficient $\phi_x^{(1)}(m)$ is defined by
\begin{align*}
\phi_x^{(1)}(m):=\frac{1}{nN}\frac{\partial G_x(m,H,\gamma)}{\partial \gamma} \Big|_{\gamma = 0}.
\end{align*}
The derivative leads to
\begin{align}
\frac{\partial G_x(m,H,\gamma)}{\partial \gamma}&=\Ave{E_x^{\mrm{int}}(\mcal{S}_x)}_{\gamma},
\label{eqn:1st-derivative_Gx}
\end{align}
where $\ave{\cdots}_{\gamma}$ denotes the average for the distribution  
\begin{align*}
P(\mcal{S}_x \mid \gamma, m) \propto \exp\big( - E_x(\mcal{S}_x;\lambda^*,\gamma)\big),
\end{align*}
where $\lambda^*$ is the value of $\lambda$ that satisfies the extremum condition in Eq.~(\ref{eqn:GibbsFreeEnergy}) 
and which is the function relating $\gamma$ and $m$; i.e., $\lambda^* = \lambda^*(\gamma,m)$. 
From the extremum condition for $\lambda$ in Eq.~(\ref{eqn:GibbsFreeEnergy}), we obtain the equation 
\begin{align}
m = \frac{1}{n\tau_x}\sum_{i=1}^n\sum_{a = 1}^{\tau_x}\ave{S_i^{\{a\}}}_{\gamma},
\label{eqn:Extr_lambda}
\end{align}
which holds for any $\gamma$.  
In the derivation of Eq.~(\ref{eqn:1st-derivative_Gx}), we used Eq.~(\ref{eqn:Extr_lambda}). 
When $\gamma = 0$, Eq.~(\ref{eqn:Extr_lambda}) reduces to $m = \tanh \lambda^*$. 
This means that $\ave{S_i^{\{a\}}}_0 = m$ for any $i$ and $a$.
Therefore, we obtain
\begin{align}
\phi_x^{(1)}(m) &=- \frac{x(n-1) N C_1}{2n} m^2 
- \frac{(n-1) K_x}{2n N}m^4,
\label{eqn:PlefkaExpansion_1st}
\end{align}
where $K_x := \tau_x(\tau_x - 1) / 2$. 
In the derivation of Eq.~(\ref{eqn:PlefkaExpansion_1st}), we used the relation 
$\ave{S_i^{\{a\}}S_j^{\{b\}}}_0 = \ave{S_i^{\{a\}}}_0\ave{S_j^{\{b\}}}_0$ if $i\not= j$ or $a \not=b$.

The coefficient $\phi_x^{(2)}(m)$ is defined by
\begin{align*}
\phi_x^{(2)}(m):=\frac{1}{2nN}\frac{\partial^2 G_x(m,H,\gamma)}{\partial \gamma^2} \Big|_{\gamma = 0}.
\end{align*}
From Eq.~(\ref{eqn:1st-derivative_Gx}), the second derivative is
\begin{align}
\frac{\partial^2 G_x(m,H,\gamma;\mcal{D})}{\partial \gamma^2}
&=\frac{\partial}{\partial \gamma}\AVE{E_x^{\mrm{int}}(\mcal{S}_x)}_{\gamma}\nn
&=\AVE{E_x^{\mrm{int}}(\mcal{S}_x)U_x(\gamma)}_{\gamma},
\label{eqn:2nd-derivative_Gx}
\end{align}
where
\begin{align*}
U_x(\gamma)&:=
\AVE{\frac{\partial E_x(\mcal{S}_x;\lambda^*,\gamma)}{\partial \gamma}}_{\gamma} 
-\frac{\partial E_x(\mcal{S}_x;\lambda^*,\gamma)}{\partial \gamma}
\end{align*}
is Georges's operator, proposed in Ref.~\cite{Georges1991}. 
To simplify the notation, we omit the explicit description of the dependency of the operator on $\mcal{S}_x$ and $m$.
By using this operator, the derivative of $\ave{A}_{\gamma}$ with respect to $\gamma$ is obtained as
\begin{align*} 
\frac{\partial \ave{A}_{\gamma}}{\partial \gamma} = \AVE{\frac{\partial A}{\partial \gamma}}_{\gamma} + \AVE{A U_x(\gamma)}_{\gamma}. 
\end{align*}
This immediately leads to $\ave{S_i^{\{a\}} U_x(\gamma)}_{\gamma} = 0$, 
because $\partial \ave{S_i^{\{a\}}}_{\gamma} /\partial \gamma = \partial m / \partial \gamma = 0$. 
Therefore, 
\begin{align}
\AVE{U_x(\gamma)^2}_{\gamma}
&=-\AVE{U_x(\gamma)\frac{\partial E_x(\mcal{S}_x,\lambda^*,\gamma)}{\partial \gamma}}_{\gamma}\nn
&=-\AVE{E_x^{\mrm{int}}(\mcal{S}_x)U_x(\gamma)}_{\gamma}
\label{eqn:Ux^2}
\end{align}
is obtained, where we have used $\ave{U_x(\gamma)}_{\gamma} = 0$.
From Eqs.~(\ref{eqn:2nd-derivative_Gx}) and (\ref{eqn:Ux^2}), we have
\begin{align}
\frac{\partial^2 G_x(m,H,\gamma)}{\partial \gamma^2}
=-\AVE{U_x(\gamma)^2}_{\gamma}.
\label{eqn:2nd-derivative_Gx_OperatorRepresentation}
\end{align}
Because
\begin{align*}
\frac{\partial \lambda^*}{\partial \gamma} \Big|_{\gamma = 0}
&= \frac{1}{n \tau_x}\frac{\partial }{\partial \gamma}\frac{\partial G_x(m,H,\gamma)}{\partial m}\Big|_{\gamma = 0}
=\frac{N}{\tau_x}\frac{\partial \phi_x^{(1)}(m)}{\partial m},
\end{align*}
when $\gamma = 0$, we obtain 
\begin{align}
&U_x(0)=\frac{(n-1)N}{n}\sum_{i=1}^n \omega_i m\sum_{a = 1}^{\tau_x}\big(S_i^{\{a\}} - m\big)\nn
&+\frac{N}{n}\sum_{i<j}\Big(d_{ij} + \frac{\tau_x - 1}{N}m^2\Big)\sum_{a = 1}^{\tau_x}\big(S_i^{\{a\}} - m\big)\big(S_j^{\{a\}} - m\big)\nn
&+\frac{1}{n}\sum_{i<j}\sum_{a<b}\big(S_i^{\{a\}}S_j^{\{a\}} - m^2\big)\big(S_i^{\{b\}}S_j^{\{b\}} - m^2\big),
\label{eqn:U0}
\end{align}
where $\omega_i$ is defined in Eq.~(\ref{eqn:def_omega_i}).

By using Eqs.~(\ref{eqn:2nd-derivative_Gx_OperatorRepresentation}) and (\ref{eqn:U0}), we obtain
\begin{widetext}
\begin{align}
\phi_x^{(2)}(m)&=-\frac{(n-1)^2  \tau_x N \Omega}{2n^2}m^2(1 - m^2)
-\frac{(n-1)\tau_x N C_2}{4n^2}(1-m^2)^2 - \frac{(n-1)K_x  C_1}{n^2}m^2(1-m^2)^2\nn
\aleq
-\frac{(n-1)K_x}{2n^2 N}\big(n + \tau_x-3\big)m^4(1-m^2)^2
-\frac{(n-1)K_x}{4n^2 N}(1 - m^4)^2,
\label{eqn:PlefkaExpansion_2nd}
\end{align}
\end{widetext}
where $\Omega$ is defined in Eq.~(\ref{eqn:def_Omega}).

\section{Evaluation of Orders of Each Term in the Empirical Bayes Likelihood}
\label{sec:app:OrdarEvaluation}

Here, we evaluate the orders of each term in Eq.~(\ref{eqn:EmpiricalBayesLikelihood_Gauss_Result}) with $m = M$, with respect to $n\gg 1$, that is, 
the orders of each term in 
\begin{align}
L_{\mrm{EB}}(H,\gamma)&\approx  e(M)-  \Phi(M)\gamma
-\phi_{-1}^{(2)}(M)\gamma^2.
\label{eqn:EmpiricalBayesLikelihood_Gauss_Result_M}
\end{align}
In the following, we assume that $N = O\big(n^{\rho} \big)$ ($\rho \geq 0$) 
and that $\{\mrm{S}_i^{(\mu)}\}$ are i.i.d. samples from a certain distribution.

First, we consider the case $H_{\mrm{true}} = 0$ in which the distribution of $\{\mrm{S}_i^{(\mu)}\}$ is unbiased. 
In this case, we obtain $M = O\big(n^{-(1+\rho)/2} \big)$, $C_1 = O\big(n^{-1-\rho/2}\big)$, and
\begin{align*}
C_2 
&= \frac{1}{N} + \frac{1}{n(n-1) N^2}\sum_{\mu < \nu} \sum_{i < j}\mrm{S}_i^{(\mu)}\mrm{S}_j^{(\mu)}\mrm{S}_i^{(\nu)}\mrm{S}_j^{(\nu)}\nn
&=O\big(n^{-\rho} \big).
\end{align*}
Similarly, we obtain
\begin{align*}
\Omega&=\frac{1}{n(n-1)^2 N^2}\sum_{i=1}^n\sum_{\mu,\nu = 1}^N \sum_{j,k \in \partial(i)}\mrm{S}_i^{(\mu)}\mrm{S}_j^{(\mu)}\mrm{S}_i^{(\nu)}\mrm{S}_k^{(\nu)}
- C_1^2\nn
&=\frac{1}{(n-1)N} 
+ \frac{1}{n(n-1)^2 N^2}\sum_{i=1}^n\sum_{\mu=1}^N \sum_{j\neq k \in \partial(i)}\mrm{S}_j^{(\mu)}\mrm{S}_k^{(\mu)}\nn
&+ \frac{1}{n(n-1)^2 N^2}\sum_{i=1}^n\sum_{\mu\neq \nu} \sum_{j,k \in \partial(i)}\mrm{S}_i^{(\mu)}\mrm{S}_j^{(\mu)}\mrm{S}_i^{(\nu)}\mrm{S}_k^{(\nu)}- C_1^2\nn
&=O\big(n^{-1-\rho} \big),
\end{align*}
because $C_1^2 = O\big( n^{-2-\rho}\big)$.
Using the above results and Eqs.~(\ref{eqn:MeanFieldEntropy}), (\ref{eqn:Phi(m)}), and (\ref{eqn:PlefkaExpansion_2nd_x=-1}), 
we obtain $e(M) = O(1)$, $\Phi(M) = O(1)$, and $\phi_{-1}^{(2)}(M) = O\big(n^{\rho-1}\big)$, respectively.  
Therefore, when $\rho = 1$, the orders of all the terms in Eq.~(\ref{eqn:EmpiricalBayesLikelihood_Gauss_Result_M}) are just $O(1)$ with respect to $n$. 

Next, we consider the case $H_{\mrm{true}} \neq 0$ in which the distribution of $\{\mrm{S}_i^{(\mu)}\}$ is biased. 
In this case, $M$, $C_1$, and $C_2$ are $O(1)$, 
and furthermore, $\Omega$ is $O(1)$ because $\omega_i = O(1)$. 
This leads to $e(M) = O(1)$, $\Phi(M) = O\big(n^{\rho}\big)$, and $\phi_{-1}^{(2)}(M) = O\big(n^{2\rho}\big)$. 
Therefore, when $\rho = 0$, the orders of all the terms in Eq.~(\ref{eqn:EmpiricalBayesLikelihood_Gauss_Result_M}) are just $O(1)$ with respect to $n$. 

This consideration and the experiments in Sec.~\ref{sec:experiment} imply that our method based on the Plefka expansion can be validated 
when all the terms in the empirical Bayes likelihood are $O(1)$.  
The introduction of the external field changes the condition to satisfy this criterion, 
leading to the appropriate scaling of $\alpha$. 
This statement is consistent with the numerical observation 
that a stable result is obtained even for different $n$'s as long as the appropriate scale in $\alpha$ is maintained, as shown in Sec.~\ref{sec:experiment}.

\subsection*{Acknowledgment}
This work was partially supported by JSPS KAKENHI (Grant Numbers: 15H03699, 18K11459, 18H03303, 25120013, and 17H00764), 
JST CREST (Grant Number: JPMJCR1402), and the COI Program from the JST (Grant Number JPMJCE1312). 
TO is also supported by a Grant for Basic Science Research Projects from the Sumitomo Foundation.  

\bibliography{citation}
\end{document}